\pdfoutput=1

\documentclass[11pt]{article}

\usepackage[preprint]{acl}

\usepackage{times}
\usepackage{latexsym}

\usepackage[T1]{fontenc}

\usepackage[utf8]{inputenc}

\usepackage{microtype}

\usepackage{inconsolata}

\usepackage{graphicx}

\usepackage{subcaption}

\usepackage[most]{tcolorbox}
\tcbuselibrary{listingsutf8}
\usepackage{listings}
\usepackage{booktabs}
\usepackage{multirow}
\usepackage{enumitem}

\title{heiDS at ArchEHR-QA 2025: \\From Fixed-$k$ to Query-dependent-$k$ for Retrieval Augmented Generation}

\author{Ashish Chouhan \and Michael Gertz \\
        Data Science Group, Institute of Computer Science \\ Heidelberg University, Germany \\ \texttt{\{chouhan, gertz\}}@informatik.uni-heidelberg.de}

\begin{document}
\maketitle
\begin{abstract}
This paper presents the approach of our team called heiDS for the ArchEHR-QA 2025 shared task. A pipeline using a retrieval augmented generation (RAG) framework is designed to generate answers that are attributed to clinical evidence from the electronic health records (EHRs) of patients  in response to patient-specific questions. We explored various components of a RAG framework, focusing  on ranked list truncation (RLT) retrieval strategies and  attribution approaches. Instead of using a fixed top-$k$ RLT retrieval strategy, we employ a query-dependent-$k$ retrieval strategy, including the existing surprise and autocut methods and two new methods proposed in this work, autocut* and elbow. The experimental results show the benefits of our strategy in producing factual and relevant answers when compared to a fixed-$k$.
\end{abstract}

\section{Introduction}
\label{sec:introduction}
Electronic Health Records (EHRs) are essential in any healthcare system, serving as repositories of the medical history of patients ~\citep{hayrinen-el-al-2008-ehr-definition}. Since 2020, patient portals have increased, resulting in more virtual communications between patients and clinicians~\citep{small-et-al-2024-large}. As a result, responding to inquiries of patients has become an important issue. Clinicians are reported to spend around 1.5 hours each day managing approximately 150 messages (patient questions)~\citep{small-et-al-2024-large, liu-et-al-2024-patient-messages}. Thus, answering patient-specific questions is a crucial task that relies on information managed in EHRs. 

Large Language Models (LLMs) can automate answer generation for patient questions, as these models are trained on extensive textual data~\citep{liu-et-al-2024-patient-messages}. However, LLMs are also prone to hallucinations, that is, they may generate answers not supported by a reliable source. This can undermine user trust and potentially harm patients by giving incorrect advice~\citep{pmlr-v235-huang24x}. Therefore, attribution, i.e., linking elements of a generated answer  to sources, is critical to ensure that every claim is grounded in medical evidence.

Attribution has gained significant attention across various domains, such as the legal and medical domains~\cite{trautmann-etal-2024-measuring, malaviya-etal-2024-expertqa}. \citet{li-el-al-2023-survey} outline three approaches for generating answers with attribution. The first approach is direct model-driven attribution, where an LLM generates answers with their sources without using  additional information. This is accomplished by fine-tuning or training the model to generate answers that include attributions~\citep{zhang-2024-longcite, patel-etal-2024-towards, huang-etal-2024-training}. However, a common issue with this approach is the hallucination of references~\citep{agrawal-etal-2024-language}. The second approach is known as post-retrieval attribution or retrieve-and-read. It retrieves evidence relevant to a query, generating an answer based on that evidence. The LLM is prompted to reference the retrieved information, thereby enforcing attribution~\citep{menick-2022-teaching, nakano-2021-webgpt, sahinuc-etal-2024-systematic, gao-etal-2023-enabling}. Post-generation attribution~\citep{gao-etal-2023-rarr, ramu-etal-2024-enhancing, cohen-wang-et-al-2024-contextcite} is the third approach, and it allows the LLM to generate answers without prior attribution and in a post-processing step map answer text back to its sources.

The objective of the BioNLP Grounded Electronic Health Record Question Answering shared task (ArchEHR-QA)~\cite{soni-etal-2025-archehr-qa} is to generate answers to patient questions, considering clinical note excerpts and attributing them with relevant evidence from the excerpts. Our approach focuses on developing a pipeline for attributed answer generation by employing a retrieval augmented generation (RAG) framework. We experimented with different methods based on the post-retrieval and post-generation attribution approaches on the ArchEHR-QA development set, which are detailed in Section~\ref{sec:sysoverview}.

\section{Pipeline Overview}
\label{sec:sysoverview}
Our proposed pipeline utilizes a RAG framework to solve the ArchEHR-QA task. This task involves answering health-related questions from patients and providing attributions based on the patients' clinical notes. In this section, we introduce our different methods, including the pipeline we submitted to the ArchEHR-QA 2025 leaderboard. Section \ref{subsec:dataset} provides information about the dataset used for the shared task, followed by Section \ref{subsec:baseline} describing the baseline. Section \ref{subsec:submittedsystem} provides information on our submitted pipeline, which is based on a surprise~\citep{bahri-et-al-2023-surprise} Ranked List Truncation (RLT) retrieval strategy.  Finally, other methods we experimented with (other than the baseline and submitted pipeline) are outlined in Section \ref{subsec:approaches}.

\subsection{Dataset}
\label{subsec:dataset}
The dataset for the ArchEHR-QA 2025 shared task, available on PhysioNet\footnote{\url{https://doi.org/10.13026/zzax-sy62} (accessed on 30th April 2025)}~\citep{soni-etal-2025-dataset-patient-needs}, comprises 20 case studies in the development (dev) set and 100 case studies in the test set\footnote{All  experiments described  in Section \ref{sec:experiments} use the dev set.}. Each case study consists of a hand-curated patient question, its corresponding clinician-rewritten version  (i.e., clinician question), and excerpts from the patient's clinical notes. See Appendix \ref{appendix:example} for an example of a  case study from the dev set and Appendix \ref{appendix:dataset_statistics} for some statistics on the clinical note excerpts. For every sentence in a clinical note, a 1024 dimensional embedding is computed using the \texttt{BAAI/bge-large-en-v1.5}\footnote{\url{https://huggingface.co/BAAI/bge-large-en-v1.5} (accessed on 4th May 2025)} model and stored in a FAISS index~\citep{johnson-et-al-2019-billion} for semantic search.

\subsection{Our Baseline}
\label{subsec:baseline}
While we experimented with various retrieval and prompting strategies within the RAG framework, our baseline follows a post-retrieval attribution approach. This involves prompting an LLM to generate answers based on both patient and clinical questions, along with \textbf{all} sentences of the clinical note excerpts from the case study. The decisions made for the baseline and other pipelines proposed in this work are supported by experiments that include
\begin{itemize}[noitemsep,topsep=0pt,parsep=0pt,partopsep=0pt]
    \item a query that is constructed using both patient and clinical questions instead of considering only one of them (see Appendix \ref{appendix:questiondecision}),
    \item a one-shot prompting approach instead of  zero-shot prompting  (see Appendix \ref{appendix:promptdecision}), 
    \item  different LLMs  for answer generation with attributions, which are \texttt{LLaMA-3.3-70B}\footnote{\url{https://huggingface.co/meta-llama/Llama-3.3-70B-Instruct} (accessed on 4th May 2025)} and \texttt{Mixtral-8x7B}\footnote{\url{https://huggingface.co/mistralai/Mixtral-8x7B-v0.1} (accessed on 4th May 2025)}~\citep{dada-etal-2025-medisumqa, kweon-et-al-2024-neurips}, and 
    \item a maximum number of 200 tokens generated by the LLM (see Appendix \ref{appendix:maxtokendecision}).
\end{itemize}

On the other hand, the organizers' baseline used the LLaMA-3.3-70B model in a zero-shot prompting approach, where the model is prompted to generate answers that include attributions. If a response is invalid, e.g., exceeding the word limit or lacking valid attribution, the model is again prompted to generate an answer. This is repeated up to five times to obtain a valid output.

\subsection{Submitted Pipeline: Surprise Ranked List Truncation (RLT) Retrieval Strategy}
\label{subsec:submittedsystem}
The pipeline we submitted for the shared task aligns with baselines utilizing a post-retrieval attribution approach. In this approach, for a query that combines patient and clinical question, semantically similar sentences from the excerpts of clinical notes are retrieved. The similarity score between the query and each sentence is computed using cosine similarity. During retrieval, $k$ represents the number of highest-scoring (top-$k$) sentences similar to the query. Instead of using a fixed value for $k$, our team employed a query-dependent-$k$ selection strategy based on the Ranked List Truncation (RLT) method, referred to as ``surprise''. This method determines the number $k$ of sentences to consider by first adjusting retrieval scores using generalized Pareto distributions from extreme value theory~\citep{piackands-1975}. It truncates a ranked list using a score threshold, allowing for a variable number of relevant sentences to be selected per query~\citep{meng-et-al-2024-rlt}. The selected sentences and query are passed to the LLMs for answer generation, where the model generates an answer with attribution explicitly referencing retrieved sentences from a clinical note. 

\begin{table*}[ht!]
\centering
\small
\caption{Retrieval performance on the development set under strict (essential only) and lenient (essential + supplementary) variants. The Strategy and Variant columns list different retrieval strategies and their parameters. Columns P, R, and F1 quantify precision, recall, and F1-score under both variants. The seven best approaches by combined strict and lenient F1-scores (excluding the $k=54$ row) are highlighted in \textbf{bold}.}
\label{tab:retrieval_performance}
\begin{tabular}{@{}l l
    c c c
    c c c@{}}
\toprule
\multirow{2}{*}{\textbf{Strategy}} & \multirow{2}{*}{\textbf{Variant}} &
  \multicolumn{3}{c}{\textbf{Strict}} &
  \multicolumn{3}{c}{\textbf{Lenient}} \\
\cmidrule(lr){3-5}\cmidrule(lr){6-8}
 &  & P & R & F1 & P & R & F1 \\
\midrule
\multirow{5}{*}{fixed-\(k\)} 
  & \(k=3\)   & 0.53 & 0.32 & 0.36 & 0.70 & 0.29 & 0.39 \\
  & \(\mathbf{k=10}\)  & 0.43 & 0.71 & \textbf{0.50} & 0.56 & 0.71 & \textbf{0.58} \\
  & \(\mathbf{k=15}\)  & 0.38 & 0.81 & \textbf{0.49} & 0.51 & 0.82 & \textbf{0.60} \\
  & \(\mathbf{k=20}\)  & 0.35 & 0.89 & \textbf{0.49} & 0.47 & 0.88 & \textbf{0.59} \\
  & \(k=54\)  & 0.33 & 1.00 & 0.49 & 0.45 & 1.00 & 0.60 \\
\addlinespace
\multirow{2}{*}{fixed-\(k\) + re-rankers}
  & \(\mathbf{FlashRank}\) (\(k=20, n=10\)) & 0.38 & 0.68 & \textbf{0.45} & 0.51 & 0.67 & \textbf{0.54} \\
  & Cohere  (\(k=20,n=10\))               & 0.38 & 0.67 & 0.45           & 0.50 & 0.66 & 0.53           \\
\midrule
autocut       & — & 0.58 & 0.22 & 0.27 & 0.68 & 0.21 & 0.28 \\
\(\mathbf{autocut*}\)      & — & 0.59 & 0.35 & \textbf{0.34} & 0.74 & 0.32 & \textbf{0.38} \\
\(\mathbf{surprise}\)      & — & 0.36 & 0.64 & \textbf{0.42} & 0.48 & 0.62 & \textbf{0.49} \\
\(\mathbf{elbow}\) & — & 0.48 & 0.66 & \textbf{0.50} & 0.62 & 0.63 & \textbf{0.55} \\
\bottomrule
\end{tabular}
\end{table*}

\subsection{Other Methods}
\label{subsec:approaches}
In this section, we outline various methods within the RAG framework by varying its components, namely retrieval strategies and attribution approaches, to assess their impact on performance. We experimented with retrieval strategies other than surprise, including fixed-$k$, fixed-$k$ and re-ranking, and query-dependent-$k$ strategies like autocut, autocut*, and elbow. 

The \textbf{Fixed-$k$} strategy applies a fixed cut-off for all query results, using common values of 3, 10, 15, 20, and 54. \textbf{Fixed-$k$ and re-ranking} is a two-step retrieval that first retrieves semantically $k$ similar candidates based on a fixed cut-off. A relevance score is assigned in the second step, selecting top-$n$ (where $n\leq k$) sentences using re-rankers like flashrank~\citep{damodaran-2023} and cohere\footnote{\url{https://docs.cohere.com/docs/rerank-overview} (accessed on 4th May 2025)}. \textbf{Autocut}\footnote{\url{https://weaviate.io/developers/weaviate/api/graphql/additional-operators\#autocut} (accessed on 4th May 2025)} limits candidate sentences based on discontinuities in the computed similarity scores. It determines the first divergence from a straight decline, excluding candidates beyond this point, although it may struggle with uniformly decreasing scores. In this work, we propose \textbf{autocut*}, a new cut-off strategy that inspects how much each similarity score decreases compared to the previous score, automatically determining cut-offs based on significant changes without any manual adjustments. We also introduce the \textbf{elbow} strategy adapted from the elbow method in clustering to determine cut-offs by plotting similarity scores and locating the ``elbow'' where the transition from high to low relevance occurs, again with no need for preset parameters.

Along with different retrieval strategies, post-generation and post-retrieval attribution approaches have also been tried. In \textbf{post‐generation attribution}, after a model generates an answer, those retrieved sentences are identified that support each answer sentence by measuring three similarity types: lexical (ROUGE~\citep{lin-2004-rouge}, BLEU~\citep{papineni-etal-2002-bleu}, METEOR~\citep{banerjee-lavie-2005-meteor}), fuzzy (character‐based matching), and semantic (BERTScore~\citep{zhang-et-al-2020-bertscore}). Each similarity is assigned a weight \(w_i\), and a combined score is calculated. If this score exceeds a predefined threshold, the candidate sentence is attributed to the generated sentence. This ensures that every claim is explicitly grounded in some original clinical evidence. Detailed setups and results on weight and threshold settings are provided in Appendix~\ref{appendix:threshold_experiments}.

The \textbf{post‐retrieval attribution} approach associates sentence identifiers with each retrieved sentence for attribution during answer generation. Post-processing steps are applied to generated answers to ensure that attributions are properly placed and no irrelevant attributions occur.

\begin{table*}[t]
\centering
\small
\caption{Pipeline evaluation on the development set under one-shot prompting, 200-token limit, and patient+clinician query.
Metrics: strict Precision (P), Recall (R), F1-score (F1), overall relevance score in \emph{Relevance} column, and overall pipeline score in \emph{Overall} column. The performance of the organizer baseline, our baseline, top three proposed pipelines, and other experimented pipelines are listed here.}
\label{tab:system_performance}
\resizebox{\textwidth}{!}{%
\begin{tabular}{l l l c c c c c}
\toprule
\textbf{Retrieval} & \textbf{Attribution} & \textbf{Model} & \textbf{P} & \textbf{R} & \textbf{F1} & \textbf{Relevance} & \textbf{Overall} \\
\midrule
\multicolumn{2}{l}{\textbf{Organizer Baseline}} & LLaMA-3.3-70B & \textbf{0.63} & \textbf{0.33} & \textbf{0.43} & 0.29 & \textbf{0.36} \\
\multicolumn{2}{l}{\textbf{Our Baseline}} & LLaMA-3.3-70B & 0.54 & 0.27 & 0.36 & 0.33 & 0.35 \\
\midrule
\multicolumn{8}{c}{\textbf{Top Three Proposed Pipelines}} \\
\midrule
surprise           & Post-retrieval   & LLaMA-3.3-70B     & 0.62 & 0.26 & 0.37 & \textbf{0.35} & \textbf{0.36} \\
elbow              & Post-retrieval   & LLaMA-3.3-70B & 0.59 & 0.27 & 0.37 & 0.32 & 0.35 \\
fixed-$k$ = 15              & Post-retrieval   & LLaMA-3.3-70B & 0.59 & 0.25 & 0.35 & 0.34 & 0.35 \\
\midrule
\multicolumn{8}{c}{\textbf{Other Experimented Pipelines}} \\
\midrule
fixed-$k$ = 10      & Post-retrieval  & LLaMA-3.3-70B   & 0.58 & 0.27 & 0.37 & 0.33 & 0.35 \\
fixed-$k$ = 10      & Post-retrieval  & Mixtral-8x7B    & 0.27 & 0.15 & 0.19 & 0.29 & 0.24 \\
fixed-$k$ = 15      & Post-retrieval  & Mixtral-8x7B    & 0.28 & 0.15 & 0.19 & 0.29 & 0.25 \\
fixed-$k$ = 20      & Post-retrieval  & LLaMA-3.3-70B   & 0.51 & 0.28 & 0.36 & 0.35 & 0.35 \\
fixed-$k$ = 20      & Post-retrieval  & Mixtral-8x7B    & 0.30 & 0.14 & 0.19 & 0.28 & 0.24 \\
fixed-$k$ = 20 + FlashRank      & Post-retrieval  & LLaMA-3.3-70B    & 0.52 & 0.22 & 0.31 & 0.34 & 0.33 \\
fixed-$k$ = 20 + FlashRank      & Post-retrieval  & Mixtral-8x7B    & 0.22 & 0.12 & 0.15 & 0.28 & 0.22 \\
autocut*             & Post-retrieval   & LLaMA-3.37B  & 0.57 & 0.14 & 0.23 & 0.32 & 0.27 \\
autocut*             & Post-retrieval   & Mixtral-8x7B & 0.44 & 0.12 & 0.18 & 0.27 & 0.23 \\
surprise             & Post-retrieval   & Mixtral-8x7B & 0.33 & 0.17 & 0.22 & 0.29 & 0.26 \\
elbow                & Post-retrieval   & Mixtral-8x7B & 0.43 & 0.15 & 0.22 & 0.29 & 0.26 \\
\midrule
fixed-$k$ = 54      & Post-generation  & LLaMA-3.3-70B & 0.35 & 0.22 & 0.27 & 0.35 & 0.31  \\
\bottomrule
\end{tabular}%
}
\end{table*}

\section{Experiments}
\label{sec:experiments}
All experiments were conducted on Google Colab\footnote{\url{https://colab.research.google.com/} (accessed on 4th May 2025)} using a Tesla T4 GPU (12GB memory)\footnote{Code for the proposed pipeline is available online: \url{https://github.com/achouhan93/heiDS-ArchEHR-QA-2025}}. For accessing LLMs, we used \texttt{InferenceClient}\footnote{\url{https://huggingface.co/docs/huggingface_hub/v0.30.2/en/package_reference/inference\_client} (accessed on 4th May 2025)} from the \texttt{huggingface\_hub} library.

\subsection{Evaluation Criteria}
\label{subsec:evaluationcriteria}
The development set provided by the organizers includes clinical note excerpts annotated with sentence numbers for attribution. Furthermore, each sentence is labeled as ``essential'', ``supplementary'', or ``not-relevant''. Evaluation is carried out for two variants, a ``strict'' variant (considering only ``essential'' labels) and a ``lenient'' variant (considering both ``essential'' and ``supplementary'' labels). Retrieval performance is measured by precision, recall, and F1-score for each variant. The results are shown in Table~\ref{tab:retrieval_performance}. We selected fixed-$k$ (10, 15, 20), autocut*, surprise, and elbow for downstream answer generation based on these metrics.

We used the official ArchEHR-QA evaluation script for the overall pipeline evaluation to assess \emph{factuality} and \emph{relevance}. \emph{Factuality} is measured by the precision, recall, and F1‐score of cited evidence versus ground‐truth annotations computed under both variants. \emph{Relevance} compares generated answer sentences to the ground‐truth essential sentences using BLEU, ROUGE, SARI~\citep{xu-etal-2016-optimizing}, BERTScore, AlignScore~\citep{zha-etal-2023-alignscore}, and MEDCON~\citep{yim2023aci}. The overall relevance score is the average of these metrics, and the final pipeline score is the mean of overall factuality (strict variant F1-score) and overall relevance.

\subsection{Comparative Pipeline Evaluation}
\label{subsec:systemevaluation}
Building on the ablations in Appendices~\ref{appendix:questiondecision}–\ref{appendix:maxtokendecision}, we fixed the query (patient + clinician question), one-shot prompt, and 200-token limit, and evaluated our pipeline with two LLMs,  \texttt{LLaMA-3.3-70B}\footnote{\url{https://huggingface.co/meta-llama/Llama-3.3-70B-Instruct} (accessed on 4th May 2025)} and \texttt{Mixtral-8x7B}\footnote{\url{https://huggingface.co/mistralai/Mixtral-8x7B-v0.1} (accessed on 4th May 2025)}, under both post-retrieval and post-generation attribution workflows. The results are shown in Table~\ref{tab:system_performance}.

\textbf{Post-Retrieval Attribution Evaluation.} We paired each of our selected retrieval strategy (fixed-$k$ = 10, 15, 20; autocut*; surprise; elbow) with each LLM and measured strict variant F1-score and overall relevance. Table~\ref{tab:system_performance} shows that \texttt{LLaMA-3.3-70B} combined with the surprise retrieval strategy achieves a strict F1-score of 0.37 and overall relevance of 0.35, making it our top post-retrieval configuration, compared to the baselines.

\textbf{Post-Generation Attribution Evaluation.} Using a fixed-$k$ of 54, we varied lexical/fuzzy/semantic weights and threshold values for the LLaMA-3.3-70B model. As shown in Table~\ref{tab:postgen_grid} in Appendix~\ref{appendix:threshold_experiments}, the optimal weighting (\(w_1=0.0, w_2=0.5, w_3=0.5\), threshold = 0.5) yields a strict F1-score of 0.27 and overall relevance score of 0.35. Although this setup performs best among post-generation configurations, it underperforms relative to the best-performing post-retrieval configuration.

\subsection{Pipeline Performance Analysis}
\label{subsec:performanceanalysis}
While our best-performing pipeline based on the surprise retrieval strategy and post-retrieval attribution achieves a comparable overall  score, it does not outperform the organizer's baseline. This outcome can be because of the following factors:
\begin{itemize}
    \item  \textbf{Prompt sensitivity of LLMs.} \citet{salinas-morstatter-2024-butterfly} demonstrate that even a small perturbation in prompts can cause changes in an LLM's output. Although the organizer baseline and our best-performing pipeline use the same model (LLaMA-3.3-70B), the organizer baseline employs a zero-shot prompt, whereas our pipeline uses a one-shot prompt with stricter formatting and attribution instructions for the model to follow. These subtle prompt design choices could have influenced the model’s ability to generate high quality answers with relevant attributions.
    \item \textbf{Difference in context size.} The development set contains up to 54 clinical note excerpt sentences per case study (see Figure~\ref{fig:dev_boxplot}), allowing the organizer baseline to input all sentences to LLM as context, thus ensuring a high recall. In contrast, our pipeline relies on a query-dependent-$k$ retrieval method to select a smaller subset of sentences. This approach naturally reduces recall, as some relevant content may not be retrieved, which thus negatively impacts the overall score.
\end{itemize}

\noindent Despite not outperforming the organizer baseline overall  score, our pipeline design is motivated by practical considerations for real-world applications. While using all clinical note sentences is feasible within the shared task environment, real-world applications can contain far more text. We consider including complete texts as  often infeasible due to LLMs input length constraints and degradation in model performance due to irrelevant information~\citep{shi-et-al-2023-irrelevant-context, liu-etal-2024-lost}. In such settings, a retrieval step is required, and determining a fixed $k$ that is suitable for all cases is time-consuming. Query-dependent-$k$ retrieval strategies remove the need for manual $k$ tuning by determining the cut-off point based on score distributions. This allows the system to adapt to different types of queries.

\section{Conclusion and Discussion}
This work explored various RAG framework components for generating answers with attributions to clinical note excerpts. Our research highlights that the best-performing pipeline employs a post-retrieval attribution approach, utilizing the ``surprise'' RLT strategy and the LLaMA-3.3-70B model. We achieved a strict variant precision of 0.62 and recall of only 0.26, resulting in an F1-score of 0.37. While this indicates that the model's attributions are often correct, it frequently overlooks relevant evidence sentences. High selectivity can be beneficial when false attributions are costly, though it may omit important information. Additionally, query-dependent-$k$ strategies like surprise, elbow, and autocut* methods for different types of queries in the dataset showed comparable performance to fixed-$k$ approaches.

\section*{Limitations}
Despite the moderate performance of our proposed pipeline, several limitations should be noted. In the current implementation, no text pre-processing is carried out for the clinical note excerpt sentences before indexing in FAISS. Expanding medical acronyms to their complete form or enriching texts with domain-specific interpretations before indexing could improve retrieval performance. Due to the use of prompting, even with a low temperature (0.001), there is non-determinism in the generated responses, making exact score replication challenging despite fixed pipeline configurations. Moreover, evaluating multiple large models increases computational requirements and associated expenses, which may limit practical deployment unless the model size or budget is adjusted.

\bibliography{anthology,custom}

\begin{thebibliography}{37}
\providecommand{\natexlab}[1]{#1}

\bibitem[{Agrawal et~al.(2024)Agrawal, Suzgun, Mackey, and
  Kalai}]{agrawal-etal-2024-language}
Ayush Agrawal, Mirac Suzgun, Lester Mackey, and Adam Kalai. 2024.
\newblock \href {https://aclanthology.org/2024.findings-eacl.62/} {Do language
  models know when they`re hallucinating references?}
\newblock In \emph{Findings of the Association for Computational Linguistics:
  EACL 2024}, pages 912--928, St. Julian{'}s, Malta. Association for
  Computational Linguistics.

\bibitem[{Bahri et~al.(2023)Bahri, Zheng, Tay, Metzler, and
  Tomkins}]{bahri-et-al-2023-surprise}
Dara Bahri, Che Zheng, Yi~Tay, Donald Metzler, and Andrew Tomkins. 2023.
\newblock \href {https://doi.org/10.1145/3539618.3592066} {Surprise: Result
  list truncation via extreme value theory}.
\newblock In \emph{Proceedings of the 46th International ACM SIGIR Conference
  on Research and Development in Information Retrieval}, SIGIR '23, page
  2404–2408. Association for Computing Machinery.

\bibitem[{Banerjee and Lavie(2005)}]{banerjee-lavie-2005-meteor}
Satanjeev Banerjee and Alon Lavie. 2005.
\newblock \href {https://aclanthology.org/W05-0909/} {{METEOR}: An automatic
  metric for {MT} evaluation with improved correlation with human judgments}.
\newblock In \emph{Proceedings of the {ACL} Workshop on Intrinsic and Extrinsic
  Evaluation Measures for Machine Translation and/or Summarization}, pages
  65--72, Ann Arbor, Michigan. Association for Computational Linguistics.

\bibitem[{Cohen-Wang et~al.(2024)Cohen-Wang, Shah, Georgiev, and
  M\k{a}dry}]{cohen-wang-et-al-2024-contextcite}
Benjamin Cohen-Wang, Harshay Shah, Kristian Georgiev, and Aleksander M\k{a}dry.
  2024.
\newblock \href
  {https://proceedings.neurips.cc/paper_files/paper/2024/file/adbea136219b64db96a9941e4249a857-Paper-Conference.pdf}
  {{ContextCite: Attributing Model Generation to Context}}.
\newblock In \emph{Advances in Neural Information Processing Systems},
  volume~37, pages 95764--95807. Curran Associates, Inc.

\bibitem[{Dada et~al.(2025)Dada, Koras, Bauer, Butler, Smith, Kleesiek, and
  Friedrich}]{dada-etal-2025-medisumqa}
Amin Dada, Osman Koras, Marie Bauer, Amanda Butler, Kaleb Smith, Jens Kleesiek,
  and Julian Friedrich. 2025.
\newblock \href {https://aclanthology.org/2025.cl4health-1.10/}
  {{M}e{D}i{S}um{QA}: Patient-oriented question-answer generation from
  discharge letters}.
\newblock In \emph{Proceedings of the Second Workshop on Patient-Oriented
  Language Processing (CL4Health)}, pages 124--136, Albuquerque, New Mexico.
  Association for Computational Linguistics.

\bibitem[{Damodaran(2023)}]{damodaran-2023}
Prithiviraj Damodaran. 2023.
\newblock \href {https://github.com/PrithivirajDamodaran/FlashRank}
  {{FlashRank, Lightest and Fastest 2nd Stage Reranker for search pipelines.}}

\bibitem[{Gao et~al.(2023{\natexlab{a}})Gao, Dai, Pasupat, Chen, Chaganty, Fan,
  Zhao, Lao, Lee, Juan, and Guu}]{gao-etal-2023-rarr}
Luyu Gao, Zhuyun Dai, Panupong Pasupat, Anthony Chen, Arun~Tejasvi Chaganty,
  Yicheng Fan, Vincent Zhao, Ni~Lao, Hongrae Lee, Da-Cheng Juan, and Kelvin
  Guu. 2023{\natexlab{a}}.
\newblock \href {https://doi.org/10.18653/v1/2023.acl-long.910} {{RARR}:
  Researching and revising what language models say, using language models}.
\newblock In \emph{Proceedings of the 61st Annual Meeting of the Association
  for Computational Linguistics (Volume 1: Long Papers)}, pages 16477--16508,
  Toronto, Canada. Association for Computational Linguistics.

\bibitem[{Gao et~al.(2023{\natexlab{b}})Gao, Yen, Yu, and
  Chen}]{gao-etal-2023-enabling}
Tianyu Gao, Howard Yen, Jiatong Yu, and Danqi Chen. 2023{\natexlab{b}}.
\newblock \href {https://doi.org/10.18653/v1/2023.emnlp-main.398} {Enabling
  large language models to generate text with citations}.
\newblock In \emph{Proceedings of the 2023 Conference on Empirical Methods in
  Natural Language Processing}, pages 6465--6488, Singapore. Association for
  Computational Linguistics.

\bibitem[{H{\"a}yrinen et~al.(2008)H{\"a}yrinen, Saranto, and
  Nyk{\"a}nen}]{hayrinen-el-al-2008-ehr-definition}
Kristiina H{\"a}yrinen, Kaija Saranto, and Pirkko Nyk{\"a}nen. 2008.
\newblock \href {https://pubmed.ncbi.nlm.nih.gov/17951106/} {Definition,
  structure, content, use and impacts of electronic health records: a review of
  the research literature}.
\newblock \emph{International journal of medical informatics}, 77(5):291--304.

\bibitem[{Huang et~al.(2024{\natexlab{a}})Huang, Wu, Hu, and
  Wang}]{huang-etal-2024-training}
Chengyu Huang, Zeqiu Wu, Yushi Hu, and Wenya Wang. 2024{\natexlab{a}}.
\newblock \href {https://doi.org/10.18653/v1/2024.acl-long.161} {Training
  language models to generate text with citations via fine-grained rewards}.
\newblock In \emph{Proceedings of the 62nd Annual Meeting of the Association
  for Computational Linguistics (Volume 1: Long Papers)}, pages 2926--2949,
  Bangkok, Thailand. Association for Computational Linguistics.

\bibitem[{Huang et~al.(2024{\natexlab{b}})Huang, Sun, Wang, Wu, Zhang, Li, Gao,
  Huang, Lyu, Zhang, Li, Sun, Liu, Liu, Wang, Zhang, Vidgen, Kailkhura, Xiong,
  Xiao, Li, Xing, Huang, Liu, Ji, Wang, Zhang, Yao, Kellis, Zitnik, Jiang,
  Bansal, Zou, Pei, Liu, Gao, Han, Zhao, Tang, Wang, Vanschoren, Mitchell, Shu,
  Xu, Chang, He, Huang, Backes, Gong, Yu, Chen, Gu, Xu, Ying, Ji, Jana, Chen,
  Liu, Zhou, Wang, Li, Zhang, Wang, Xie, Chen, Wang, Liu, Ye, Cao, Chen, and
  Zhao}]{pmlr-v235-huang24x}
Yue Huang, Lichao Sun, Haoran Wang, Siyuan Wu, Qihui Zhang, Yuan Li, Chujie
  Gao, Yixin Huang, Wenhan Lyu, Yixuan Zhang, Xiner Li, Hanchi Sun, Zhengliang
  Liu, Yixin Liu, Yijue Wang, Zhikun Zhang, Bertie Vidgen, Bhavya Kailkhura,
  Caiming Xiong, and 52 others. 2024{\natexlab{b}}.
\newblock \href {https://proceedings.mlr.press/v235/huang24x.html} {Position:
  {T}rust{LLM}: Trustworthiness in large language models}.
\newblock In \emph{Proceedings of the 41st International Conference on Machine
  Learning}, volume 235 of \emph{Proceedings of Machine Learning Research},
  pages 20166--20270. PMLR.

\bibitem[{Johnson et~al.(2019)Johnson, Douze, and
  J{\'e}gou}]{johnson-et-al-2019-billion}
Jeff Johnson, Matthijs Douze, and Herv{\'e} J{\'e}gou. 2019.
\newblock \href {https://ieeexplore.ieee.org/document/8733051} {Billion-scale
  similarity search with {GPUs}}.
\newblock \emph{IEEE Transactions on Big Data}, 7(3):535--547.

\bibitem[{Kweon et~al.(2024)Kweon, Kim, Kwak, Cha, Yoon, Kim, Yang, Won, and
  Choi}]{kweon-et-al-2024-neurips}
Sunjun Kweon, Jiyoun Kim, Heeyoung Kwak, Dongchul Cha, Hangyul Yoon, Kwanghyun
  Kim, Jeewon Yang, Seunghyun Won, and Edward Choi. 2024.
\newblock \href {https://nips.cc/virtual/2024/poster/97643} {{EHRNoteQA: An LLM
  Benchmark for Real-World Clinical Practice Using Discharge Summaries}}.
\newblock In \emph{Advances in Neural Information Processing Systems},
  volume~37, pages 124575--124611. Curran Associates, Inc.

\bibitem[{Li et~al.(2023)Li, Sun, Hu, Liu, Chen, Hu, Wu, and
  Zhang}]{li-el-al-2023-survey}
Dongfang Li, Zetian Sun, Xinshuo Hu, Zhenyu Liu, Ziyang Chen, Baotian Hu, Aiguo
  Wu, and Min Zhang. 2023.
\newblock \href {https://arxiv.org/abs/2311.03731} {{A Survey of Large Language
  Models Attribution}}.
\newblock \emph{arXiv preprint arXiv:2311.03731}.

\bibitem[{Lin(2004)}]{lin-2004-rouge}
Chin-Yew Lin. 2004.
\newblock \href {https://aclanthology.org/W04-1013/} {{ROUGE}: A package for
  automatic evaluation of summaries}.
\newblock In \emph{Text Summarization Branches Out}, pages 74--81, Barcelona,
  Spain. Association for Computational Linguistics.

\bibitem[{Liu et~al.(2024{\natexlab{a}})Liu, Lin, Hewitt, Paranjape,
  Bevilacqua, Petroni, and Liang}]{liu-etal-2024-lost}
Nelson~F. Liu, Kevin Lin, John Hewitt, Ashwin Paranjape, Michele Bevilacqua,
  Fabio Petroni, and Percy Liang. 2024{\natexlab{a}}.
\newblock \href {https://doi.org/10.1162/tacl_a_00638} {Lost in the middle: How
  language models use long contexts}.
\newblock \emph{Transactions of the Association for Computational Linguistics},
  12:157--173.

\bibitem[{Liu et~al.(2024{\natexlab{b}})Liu, McCoy, Wright, Carew, Genkins,
  Huang, Peterson, Steitz, and Wright}]{liu-et-al-2024-patient-messages}
Siru Liu, Allison~B McCoy, Aileen~P Wright, Babatunde Carew, Julian~Z Genkins,
  Sean~S Huang, Josh~F Peterson, Bryan Steitz, and Adam Wright.
  2024{\natexlab{b}}.
\newblock \href {https://pubmed.ncbi.nlm.nih.gov/37503263/} {{Leveraging large
  language models for generating responses to patient messages—a subjective
  analysis}}.
\newblock \emph{Journal of the American Medical Informatics Association},
  31(6):1367--1379.

\bibitem[{Malaviya et~al.(2024)Malaviya, Lee, Chen, Sieber, Yatskar, and
  Roth}]{malaviya-etal-2024-expertqa}
Chaitanya Malaviya, Subin Lee, Sihao Chen, Elizabeth Sieber, Mark Yatskar, and
  Dan Roth. 2024.
\newblock \href {https://doi.org/10.18653/v1/2024.naacl-long.167}
  {{E}xpert{QA}: Expert-curated questions and attributed answers}.
\newblock In \emph{Proceedings of the 2024 Conference of the North American
  Chapter of the Association for Computational Linguistics: Human Language
  Technologies (Volume 1: Long Papers)}, pages 3025--3045, Mexico City, Mexico.
  Association for Computational Linguistics.

\bibitem[{Meng et~al.(2024)Meng, Arabzadeh, Askari, Aliannejadi, and
  de~Rijke}]{meng-et-al-2024-rlt}
Chuan Meng, Negar Arabzadeh, Arian Askari, Mohammad Aliannejadi, and Maarten
  de~Rijke. 2024.
\newblock \href {https://dl.acm.org/doi/10.1145/3626772.3657864} {{Ranked List
  Truncation for Large Language Model-based Re-Ranking}}.
\newblock In \emph{Proceedings of the 47th International ACM SIGIR Conference
  on Research and Development in Information Retrieval}, SIGIR '24, page
  141–151, New York, NY, USA. Association for Computing Machinery.

\bibitem[{Menick et~al.(2022)Menick, Trebacz, Mikulik, Aslanides, Song,
  Chadwick, Glaese, Young, Campbell-Gillingham, Irving
  et~al.}]{menick-2022-teaching}
Jacob Menick, Maja Trebacz, Vladimir Mikulik, John Aslanides, Francis Song,
  Martin Chadwick, Mia Glaese, Susannah Young, Lucy Campbell-Gillingham,
  Geoffrey Irving, and 1 others. 2022.
\newblock \href {https://arxiv.org/abs/2203.11147} {{Teaching language models
  to support answers with verified quotes}}.
\newblock \emph{arXiv preprint arXiv:2203.11147}.

\bibitem[{Nakano et~al.(2021)Nakano, Hilton, Balaji, Wu, Ouyang, Kim, Hesse,
  Jain, Kosaraju, Saunders et~al.}]{nakano-2021-webgpt}
Reiichiro Nakano, Jacob Hilton, Suchir Balaji, Jeff Wu, Long Ouyang, Christina
  Kim, Christopher Hesse, Shantanu Jain, Vineet Kosaraju, William Saunders, and
  1 others. 2021.
\newblock \href {https://arxiv.org/abs/2112.09332} {{WebGPT: Browser-assisted
  question-answering with human feedback}}.
\newblock \emph{arXiv preprint arXiv:2112.09332}.

\bibitem[{Papineni et~al.(2002)Papineni, Roukos, Ward, and
  Zhu}]{papineni-etal-2002-bleu}
Kishore Papineni, Salim Roukos, Todd Ward, and Wei-Jing Zhu. 2002.
\newblock \href {https://doi.org/10.3115/1073083.1073135} {{B}leu: a method for
  automatic evaluation of machine translation}.
\newblock In \emph{Proceedings of the 40th Annual Meeting of the Association
  for Computational Linguistics}, pages 311--318, Philadelphia, Pennsylvania,
  USA. Association for Computational Linguistics.

\bibitem[{Patel et~al.(2024)Patel, Subramanian, Garg, Banerjee, and
  Misra}]{patel-etal-2024-towards}
Nilay Patel, Shivashankar Subramanian, Siddhant Garg, Pratyay Banerjee, and
  Amita Misra. 2024.
\newblock \href {https://doi.org/10.18653/v1/2024.naacl-long.216} {Towards
  improved multi-source attribution for long-form answer generation}.
\newblock In \emph{Proceedings of the 2024 Conference of the North American
  Chapter of the Association for Computational Linguistics: Human Language
  Technologies (Volume 1: Long Papers)}, pages 3906--3919, Mexico City, Mexico.
  Association for Computational Linguistics.

\bibitem[{Pickands(1975)}]{piackands-1975}
James Pickands. 1975.
\newblock \href
  {https://projecteuclid.org/journals/annals-of-statistics/volume-3/issue-1/Statistical-Inference-Using-Extreme-Order-Statistics/10.1214/aos/1176343003.full}
  {Statistical inference using extreme order statistics}.
\newblock \emph{The Annals of Statistics}, 3(1):119--131.

\bibitem[{Ramu et~al.(2024)Ramu, Goswami, Saxena, and
  Srinivasan}]{ramu-etal-2024-enhancing}
Pritika Ramu, Koustava Goswami, Apoorv Saxena, and Balaji~Vasan Srinivasan.
  2024.
\newblock \href {https://doi.org/10.18653/v1/2024.emnlp-main.985} {Enhancing
  post-hoc attributions in long document comprehension via coarse grained
  answer decomposition}.
\newblock In \emph{Proceedings of the 2024 Conference on Empirical Methods in
  Natural Language Processing}, pages 17790--17806, Miami, Florida, USA.
  Association for Computational Linguistics.

\bibitem[{{\c{S}}ahinu{\c{c}} et~al.(2024){\c{S}}ahinu{\c{c}}, Kuznetsov, Hou,
  and Gurevych}]{sahinuc-etal-2024-systematic}
Furkan {\c{S}}ahinu{\c{c}}, Ilia Kuznetsov, Yufang Hou, and Iryna Gurevych.
  2024.
\newblock \href {https://doi.org/10.18653/v1/2024.acl-long.265} {Systematic
  task exploration with {LLM}s: A study in citation text generation}.
\newblock In \emph{Proceedings of the 62nd Annual Meeting of the Association
  for Computational Linguistics (Volume 1: Long Papers)}, pages 4832--4855,
  Bangkok, Thailand. Association for Computational Linguistics.

\bibitem[{Salinas and Morstatter(2024)}]{salinas-morstatter-2024-butterfly}
Abel Salinas and Fred Morstatter. 2024.
\newblock \href {https://doi.org/10.18653/v1/2024.findings-acl.275} {The
  butterfly effect of altering prompts: How small changes and jailbreaks affect
  large language model performance}.
\newblock In \emph{Findings of the Association for Computational Linguistics:
  ACL 2024}, pages 4629--4651, Bangkok, Thailand. Association for Computational
  Linguistics.

\bibitem[{Shi et~al.(2023)Shi, Chen, Misra, Scales, Dohan, Chi, Sch\"{a}rli,
  and Zhou}]{shi-et-al-2023-irrelevant-context}
Freda Shi, Xinyun Chen, Kanishka Misra, Nathan Scales, David Dohan, Ed~H. Chi,
  Nathanael Sch\"{a}rli, and Denny Zhou. 2023.
\newblock Large language models can be easily distracted by irrelevant context.
\newblock In \emph{Proceedings of the 40th International Conference on Machine
  Learning}, volume 202 of \emph{Proceedings of Machine Learning Research},
  pages 31210--31227. PMLR.

\bibitem[{Small et~al.(2024)Small, Wiesenfeld, Brandfield-Harvey, Jonassen,
  Mandal, Stevens, Major, Lostraglio, Szerencsy, Jones
  et~al.}]{small-et-al-2024-large}
William~R Small, Batia Wiesenfeld, Beatrix Brandfield-Harvey, Zoe Jonassen,
  Soumik Mandal, Elizabeth~R Stevens, Vincent~J Major, Erin Lostraglio, Adam
  Szerencsy, Simon Jones, and 1 others. 2024.
\newblock \href {https://pubmed.ncbi.nlm.nih.gov/39012633/} {{Large Language
  Model--Based Responses to Patients’ In-Basket Messages}}.
\newblock \emph{JAMA network open}, 7(7):e2422399--e2422399.

\bibitem[{Soni and
  Demner-Fushman(2025{\natexlab{a}})}]{soni-etal-2025-dataset-patient-needs}
Sarvesh Soni and Dina Demner-Fushman. 2025{\natexlab{a}}.
\newblock {A Dataset for Addressing Patient's Information Needs related to
  Clinical Course of Hospitalization}.
\newblock \emph{arXiv preprint}.

\bibitem[{Soni and
  Demner-Fushman(2025{\natexlab{b}})}]{soni-etal-2025-archehr-qa}
Sarvesh Soni and Dina Demner-Fushman. 2025{\natexlab{b}}.
\newblock {Overview of the ArchEHR-QA 2025 Shared Task on Grounded Question
  Answering from Electronic Health Records}.
\newblock In \emph{The 24th Workshop on Biomedical Natural Language Processing
  and BioNLP Shared Tasks}, Vienna, Austria. Association for Computational
  Linguistics.

\bibitem[{Trautmann et~al.(2024)Trautmann, Ostapuk, Grail, Pol, Bonifazi, Gao,
  and Gajek}]{trautmann-etal-2024-measuring}
Dietrich Trautmann, Natalia Ostapuk, Quentin Grail, Adrian Pol, Guglielmo
  Bonifazi, Shang Gao, and Martin Gajek. 2024.
\newblock \href {https://doi.org/10.18653/v1/2024.nllp-1.14} {Measuring the
  groundedness of legal question-answering systems}.
\newblock In \emph{Proceedings of the Natural Legal Language Processing
  Workshop 2024}, pages 176--186, Miami, FL, USA. Association for Computational
  Linguistics.

\bibitem[{Xu et~al.(2016)Xu, Napoles, Pavlick, Chen, and
  Callison-Burch}]{xu-etal-2016-optimizing}
Wei Xu, Courtney Napoles, Ellie Pavlick, Quanze Chen, and Chris Callison-Burch.
  2016.
\newblock \href {https://doi.org/10.1162/tacl_a_00107} {Optimizing statistical
  machine translation for text simplification}.
\newblock \emph{Transactions of the Association for Computational Linguistics},
  4:401--415.

\bibitem[{Yim et~al.(2023)Yim, Fu, Ben~Abacha, Snider, Lin, and
  Yetisgen}]{yim2023aci}
Wen-wai Yim, Yujuan Fu, Asma Ben~Abacha, Neal Snider, Thomas Lin, and Meliha
  Yetisgen. 2023.
\newblock \href {https://www.nature.com/articles/s41597-023-02487-3}
  {{Aci-bench: a Novel Ambient Clinical Intelligence Dataset for Benchmarking
  Automatic Visit Note Generation}}.
\newblock \emph{Scientific data}, 10(1):586.

\bibitem[{Zha et~al.(2023)Zha, Yang, Li, and Hu}]{zha-etal-2023-alignscore}
Yuheng Zha, Yichi Yang, Ruichen Li, and Zhiting Hu. 2023.
\newblock \href {https://doi.org/10.18653/v1/2023.acl-long.634}
  {{A}lign{S}core: Evaluating factual consistency with a unified alignment
  function}.
\newblock In \emph{Proceedings of the 61st Annual Meeting of the Association
  for Computational Linguistics (Volume 1: Long Papers)}, pages 11328--11348,
  Toronto, Canada. Association for Computational Linguistics.

\bibitem[{Zhang et~al.(2024)Zhang, Bai, Lv, Gu, Liu, Zou, Cao, Hou, Dong, Feng
  et~al.}]{zhang-2024-longcite}
Jiajie Zhang, Yushi Bai, Xin Lv, Wanjun Gu, Danqing Liu, Minhao Zou, Shulin
  Cao, Lei Hou, Yuxiao Dong, Ling Feng, and 1 others. 2024.
\newblock \href {https://arxiv.org/abs/2409.02897} {{LongCite: Enabling LLMs to
  Generate Fine-grained Citations in Long-context QA}}.
\newblock \emph{arXiv preprint arXiv:2409.02897}.

\bibitem[{Zhang et~al.(2020)Zhang, Kishore, Wu, Weinberger, and
  Artzi}]{zhang-et-al-2020-bertscore}
Tianyi Zhang, Varsha Kishore, Felix Wu, Kilian~Q. Weinberger, and Yoav Artzi.
  2020.
\newblock \href {https://arxiv.org/abs/1904.09675} {{BERTScore: Evaluating Text
  Generation with {BERT}}}.
\newblock In \emph{Proceedings of the Eighth International Conference on
  Learning Representations ({ICLR}'20)}. OpenReview.net.

\end{thebibliography}

\appendix
\section{Example Case Study}
\label{appendix:example}

\begin{tcolorbox}[colback=gray!3, colframe=black!60, title=Example Case: Patient and Clinician Questions with Clinical Note, boxrule=0.4pt, arc=2pt, left=5pt, right=5pt, top=4pt, bottom=4pt]

\textbf{Patient Question}: \\
Took my 59 yo father to ER ultrasound discovered he had an aortic aneurysm. He had a salvage repair (tube graft). Long surgery / recovery for couple hours then removed packs. why did they do this surgery????? After this time he spent 1 month in hospital now sent home.

\vspace{1.5ex}
\textbf{Clinician Question}: \\
Why did they perform the emergency salvage repair on him?

\vspace{1.5ex}
\textbf{Clinical Note}:
\textbf{1:} He was transferred to the hospital on 2025-1-20 for emergent repair of his ruptured thoracoabdominal aortic aneurysm. 
\textbf{2:} He was immediately taken to the operating room where he underwent an emergent salvage repair of ruptured thoracoabdominal aortic aneurysm with a 34-mm Dacron tube graft using deep hypothermic circulatory arrest. 
\textbf{3:} Please see operative note for details which included cardiac arrest x2. 
\textbf{4:} Postoperatively he was taken to the intensive care unit for monitoring with an open chest. 
\textbf{5:} He remained intubated and sedated on pressors and inotropes. 
\textbf{6:} On 2025-1-22, he returned to the operating room where he underwent exploration and chest closure. 
\textbf{7:} On 1-25 he returned to the OR for abdominal closure, JP drain placement, and feeding jejunostomy placed at that time for nutritional support. 
\textbf{8:} Thoracoabdominal wound healing well with exception of very small open area mid-wound that is approximately 1cm around and 0.5cm deep, with no surrounding erythema. 
\textbf{9:} Packed with dry gauze and covered with DSD.

\end{tcolorbox}

\section{Dataset Statistics}
\label{appendix:dataset_statistics}
\begin{figure}[t]
    \centering

    \begin{subfigure}[t]{\linewidth}
        \centering
        \includegraphics[width=\linewidth]{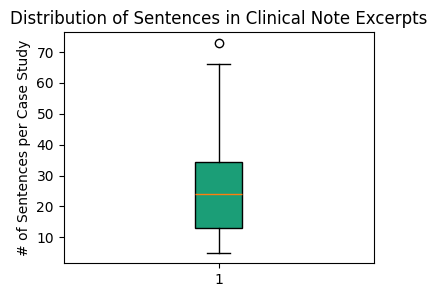}
        \caption{Test set}
        \label{fig:test_boxplot}
    \end{subfigure}

    \vspace{4mm}  

    \begin{subfigure}[t]{\linewidth}
        \centering
        \includegraphics[width=\linewidth]{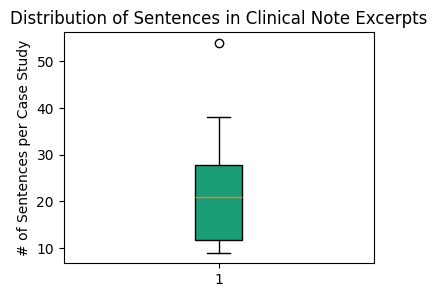}
        \caption{Development set}
        \label{fig:dev_boxplot}
    \end{subfigure}

    \caption{Distribution of the number of sentences per clinical case in the test (a) and development (b) sets.}
    \label{fig:sentence_distribution_boxplots}
\end{figure}

The box plots representing the distribution of sentences in clinical notes for development (dev) and test sets (see Figure \ref{fig:test_boxplot} and \ref{fig:dev_boxplot}) show that there is a varying  number of sentences present in different case studies with outliers (in the dev set case study, No.~8 is having 54 sentences, and in the test set case study, No.~73 is having 74 sentences). 

\begin{figure}[t]
    \centering

    \begin{subfigure}[t]{\linewidth}
        \centering
        \includegraphics[width=\linewidth]{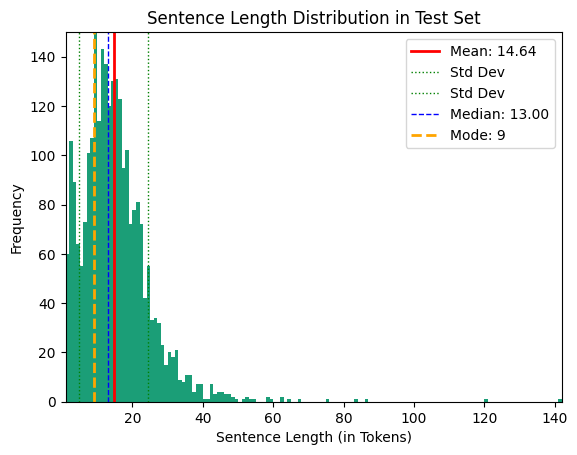}
        \caption{Test set}
        \label{fig:test_sentence_distribution}
    \end{subfigure}

    \vspace{4mm}  

    \begin{subfigure}[t]{\linewidth}
        \centering
        \includegraphics[width=\linewidth]{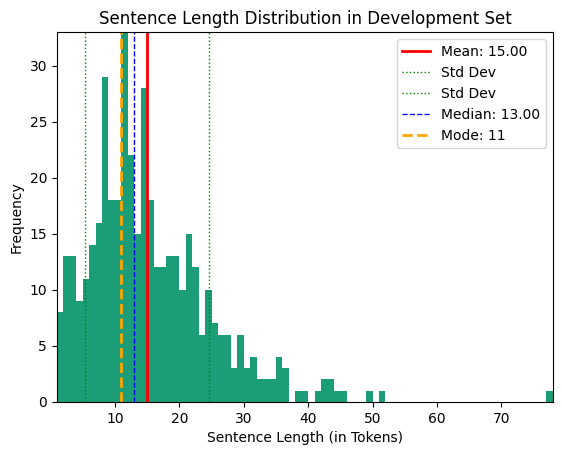}
        \caption{Development set}
        \label{fig:devlopment_sentence_distribution}
    \end{subfigure}

    \caption{Distribution of the sentence length in the test (a) and development (b) sets.}
    \label{fig:sentence_length_distribution}
\end{figure}

Similarly, when the sentence length distributions are plotted for the dev set and the test set (see Figure \ref{fig:test_sentence_distribution} and Figure \ref{fig:devlopment_sentence_distribution}), the mean of sentence length for both is nearly the same, around 15. However, in the test set, case studies have sentences that are double the length of sentences present in the dev set. 

\section{Prompt Templates for Clinical Answer Generation}
\label{appendix:prompts}

In this section, we provide the prompt templates used for answer generation. Zero-shot and one-shot prompts are designed for both post-retrieval and post-generation attribution. Post-retrieval attribution guides the language model to generate answers with inline citations, whereas post-generation attribution focuses only on answer generation, followed by a separate attribution step.

\vspace{1em}
\subsection{Prompt 1}
\label{prompt:postret_zs}
\begin{tcolorbox}[title=Zero-Shot Prompting for Post-Retrieval Attribution Approach, colback=gray!3, colframe=black!60, fonttitle=\bfseries,
breakable, listing only, listing options={basicstyle=\scriptsize\ttfamily, breaklines=true, columns=fullflexible}]
You are a clinical response generation system responsible for producing answers to health-related questions using the provided clinical note excerpts. Your answer MUST be:\\
- **Accurate and Factual:** Grounded STRICTLY in the provided clinical note excerpts ONLY.\\
- **Neutral and Objective:** DO NOT INCLUDE PERSONAL OPINIONS, NOTES, IRRELEVANT, OR UNRELATED comments.\\
- **Concise and Relevant:** INCLUDE only clinically supported statements using the exact terminology found in the provided clinical notes. Do not add any additional interpretations or synonyms.\\
- **Third-Person Perspective:** Do not address the reader directly.\\
- **Citation:** Each statement must be supported by a NUMBERED CLINICAL NOTE SENTENCE from the Clinical Note Excerpts ONLY. The citation must be placed strictly AT THE END of the sentence. DO NOT insert citations within the sentence or phrase. When citing a single source, cite it as |id|. When a statement is supported by multiple sources, combine their IDs within a single pair of vertical bars (e.g., |id, id, id|) with IDs separated by commas and no extra vertical bars.\\
- **Mandatory Citation Inclusion:** AT LEAST ONE SENTENCE in your answer MUST include a citation from the provided clinical notes.\\
\\
**Inputs:**\\
1. **Clinical Note Excerpts:** Retrieved sentences from the patient's clinical record, numbered.\\
2. **Patient Narrative Context:** Additional context from the patient's perspective.\\
3. **Clinician Question:** The primary question requiring an answer.\\
\\
**Your Task:**\\
Generate a response based strictly on the provided input. Follow the structured format exactly, use only the exact terms from the clinical note excerpts, and ensure all citations are formatted consistently.\\

[Clinical Note Begin]\newline
\{note\}\newline
[Clinical Note End]\newline
\newline
[Patient Narrative Context Begin]\newline
\{patient\_narrative\}\newline
[Patient Narrative Context End]\newline
\newline
[Clinician Question Begin]\newline
\{clinician\_question\}\newline
[Clinician Question End]\newline

Provide your structured answer below:
\end{tcolorbox}

\vspace{1em}
\subsection{Prompt 2}
\label{prompt:postret_os}

\begin{tcolorbox}[title=One-Shot Prompting for Post-Retrieval Attribution Approach, colback=gray!3, colframe=black!60, fonttitle=\bfseries,
breakable, listing only, listing options={basicstyle=\scriptsize\ttfamily, breaklines=true, columns=fullflexible}]
You are a clinical response generation system responsible for producing answers to health-related questions ... \newline
[ ... TRUNCATED FOR BREVITY ... ] \newline

**Example:**\newline
If the clinician asks, "Why did they perform the emergency salvage repair on him?", and the note states:\newline
1: He was transferred to the hospital on 2025-1-20 for emergent repair of his ruptured thoracoabdominal aortic aneurysm.\newline
2: He was immediately taken to the operating room where he underwent an emergent salvage repair of ruptured thoracoabdominal aortic aneurysm with a 34-mm Dacron tube graft using deep hypothermic circulatory arrest.\newline
Then the response should be:\newline
His aortic aneurysm was caused by the rupture of a thoracoabdominal aortic aneurysm, which required emergent surgical intervention |1|. He underwent a complex salvage repair using a 34-mm Dacron tube graft and deep hypothermic circulatory arrest to address the rupture |2|.\newline
[ ... TRUNCATED FOR BREVITY ... ] \newline

Provide your structured answer below:
\end{tcolorbox}

\vspace{1em}

\subsection{Prompt 3}
\label{prompt:postgen_zs}
\begin{tcolorbox}[title=Zero-Shot Prompting for Post-Generation Attribution Approach, colback=gray!3, colframe=black!60, fonttitle=\bfseries,
breakable, listing only, listing options={basicstyle=\scriptsize\ttfamily, breaklines=true, columns=fullflexible}]
You are a clinical response generation system responsible for producing answers to health-related questions using the provided clinical note excerpts. Your answer MUST be:\newline
- **Accurate and Factual:** Grounded STRICTLY in the provided clinical note excerpts ONLY.\newline
- **Neutral and Objective:** DO NOT INCLUDE PERSONAL OPINIONS, NOTES, IRRELEVANT, OR UNRELATED comments.\newline
- **Concise and Relevant:** INCLUDE only clinically supported statements using the exact terminology found in the provided clinical notes. Do not add any additional interpretations or synonyms.\newline
- **Third-Person Perspective:** Do not address the reader directly."\newline

**Inputs:**\\
1. **Clinical Note Excerpts:** Retrieved sentences from the patient's clinical record, numbered.\\
2. **Patient Narrative Context:** Additional context from the patient's perspective.\\
3. **Clinician Question:** The primary question requiring an answer.\\
\\
**Your Task:**\\
Generate a response based strictly on the provided input. Follow the structured format exactly, use only the exact terms from the clinical note excerpts, and ensure all citations are formatted consistently.\\

[Clinical Note Begin]\newline
\{note\}\newline
[Clinical Note End]\newline
\newline
[Patient Narrative Context Begin]\newline
\{patient\_narrative\}\newline
[Patient Narrative Context End]\newline
\newline
[Clinician Question Begin]\newline
\{clinician\_question\}\newline
[Clinician Question End]\newline

Provide your structured answer below:
\end{tcolorbox}

\vspace{1em}

\subsection{Prompt 4}
\label{prompt:postgen_os}
\begin{tcolorbox}[title=One-Shot Prompting for Post-Generation Attribution Approach, colback=gray!3, colframe=black!60, fonttitle=\bfseries,
breakable, listing only, listing options={basicstyle=\scriptsize\ttfamily, breaklines=true, columns=fullflexible}]
You are a clinical response generation system responsible for producing answers to health-related questions ... \newline
[ ... TRUNCATED FOR BREVITY ... ] \newline

**Example:**\newline
If the clinician asks, "Why did they perform the emergency salvage repair on him?", and the note states:\newline
1: He was transferred to the hospital on 2025-1-20 for emergent repair of his ruptured thoracoabdominal aortic aneurysm.\newline
2: He was immediately taken to the operating room where he underwent an emergent salvage repair of ruptured thoracoabdominal aortic aneurysm with a 34-mm Dacron tube graft using deep hypothermic circulatory arrest.\newline
Then the response should be:\newline
His aortic aneurysm was caused by the rupture of a thoracoabdominal aortic aneurysm, which required emergent surgical intervention. He underwent a complex salvage repair using a 34-mm Dacron tube graft and deep hypothermic circulatory arrest to address the rupture.\newline
[ ... TRUNCATED FOR BREVITY ... ] \newline

Provide your structured answer below:
\end{tcolorbox}

\section{Query Formulation Experiment}
\label{appendix:questiondecision}
We compared three query formulation approaches. First, the patient's question is used; second, the clinician's question is used; third, both patient and clinician questions are considered. The setup for an experiment is similar to the baseline (see Section \ref{subsec:baseline}), i.e., \textbf{all} clinical notes excerpt sentences for each case study are considered and passed to LLaMA-3.3-70B (initially, the configuration is set to a maximum token generation of 100 tokens and zero-shot prompting). Table~\ref{tab:query_decision} shows the overall factuality (strict variant F1-score), relevance, and pipeline scores, demonstrating that combining patient and clinician questions yields the best performance.

\begin{table}[ht]
\centering
\small
\caption{Query Formulation Results. All experiments use a fixed-\(k=54\), zero-shot prompting, post-retrieval attribution with \texttt{LLaMA-3.3-70B} model, and a maximum token limit of 100. Metrics: strict Precision (P), strict F1 (F1), overall Relevance (R), and overall pipeline score (O). The best variant is highlighted in \textbf{bold}.}
\label{tab:query_decision}
\resizebox{\columnwidth}{!}{%
\begin{tabular}{lcccc}
\toprule
\textbf{Query} & \textbf{P} & \textbf{F1} & \textbf{R} & \textbf{O} \\
\midrule
Patient Question only            & 0.39 & 0.27 & 0.33 & 0.30 \\
Clinician Question only          & 0.42 & 0.27 & 0.30 & 0.28 \\
\(\mathbf{Patient+Clinician}\) & \textbf{0.44} & \textbf{0.30} & \textbf{0.33} & \textbf{0.31} \\
\bottomrule
\end{tabular}%
}
\end{table}

\section{Prompting Approach Experiment}
\label{appendix:promptdecision}
To assess the effect of the prompting approach, we compared zero-shot and one-shot prompting approaches considering the LLaMA-3.3-70B model and prompting with \textbf{all} note sentences, the query as a combination of patient and clinician questions (see Appendix~\ref{appendix:questiondecision}), and a maximum token generation limit of 100. LLMs generate answers based solely on the provided query and instructions in a zero-shot prompting approach, testing their inherent understanding without examples. See Appendices \ref{prompt:postret_zs} and \ref{prompt:postgen_zs} for zero-shot prompts. In the one-shot prompting approach, an example of the desired output is provided alongside the query and instructions, helping the model align its response style. See Appendices \ref{prompt:postret_os} and \ref{prompt:postgen_os} for one-shot prompts. Table~\ref{tab:prompt_decision} shows the overall factuality (strict variant F1-score), relevance, and pipeline score for each approach. The one-shot prompt yielded higher scores, leading us to select it for the baseline and methods.

\begin{table}[ht]
\centering
\small
\caption{Prompting Approach Results. All experiments use fixed-\(k=54\), query (patient + clinical questions), post-retrieval attribution with \texttt{LLaMA-3.3-70B} model, and a maximum token limit of 100. Metrics: strict Precision (P), strict F1 (F1), overall Relevance (R), and overall pipeline score (O). The best variant is highlighted in \textbf{bold}.}
\label{tab:prompt_decision}
\resizebox{\columnwidth}{!}{%
\begin{tabular}{lcccc}
\toprule
\textbf{Prompting Approach} & \textbf{P} & \textbf{F1} & \textbf{R} & \textbf{O} \\
\midrule
zero-shot prompting            & 0.44 & 0.30 & 0.33 & 0.31 \\
\(\textbf{one-shot prompting}\) & \textbf{0.56} & \textbf{0.34} & \textbf{0.33} & \textbf{0.33} \\
\bottomrule
\end{tabular}%
}
\end{table}

\begin{table*}[t]
\centering
\caption{Parameter Settings. Experiments use fixed-\(k=54\), query (patient+clinician question), one-shot prompting, and post-generation attribution with \texttt{LLaMA-3.3-70B} model. Metrics: strict Precision (P), strict F1 (F1), overall Relevance (R), and overall pipeline score (O). Different combinations of weights and thresholds are arranged in descending order of performance, i.e., the best combination at the top.}
\label{tab:postgen_grid}
\begin{tabular}{cccccccc}
\toprule
$w_{1}$ & $w_{2}$ & $w_{3}$ & $T$ & P & F1 & R & Overall \\
\midrule
\textbf{0.0} & \textbf{0.5} & \textbf{0.5} & \textbf{0.5} & \textbf{0.35} & \textbf{0.27} & \textbf{0.35} & \textbf{0.311} \\
0.3 & 0.4 & 0.3 & 0.4 & 0.34 & 0.27 & 0.35 & 0.307 \\
0.3 & 0.3 & 0.4 & 0.4 & 0.32 & 0.26 & 0.35 & 0.306 \\
0.2 & 0.4 & 0.4 & 0.4 & 0.28 & 0.25 & 0.35 & 0.300 \\
0.5 & 0.5 & 0.0 & 0.3 & 0.30 & 0.26 & 0.34 & 0.300 \\
\bottomrule
\end{tabular}
\end{table*}

\section{Maximum Token Generation Experiment}
\label{appendix:maxtokendecision}
We experimented with the LLaMA-3.3-70B model having maximum token generation limits of 100, 200, and 300 tokens\footnote{Approximately corresponding to the organizer’s 75-word guideline.} to determine their impact on the pipeline's overall performance. Table~\ref{tab:token_decision} shows that a maximum number of 200 tokens achieved the best balance of overall factuality (strict variant F1-score) and relevance scores. Consequently, we fixed the maximum number of tokens to 200  in all experiments.

\begin{table}[ht]
\centering
\small
\caption{Maximum Token Generation. All experiments use fixed-\(k=54\), query (patient+clinician question), one-shot prompting, and post-retrieval attribution with \texttt{LLaMA-3.3-70B} model. Metrics: strict Precision (P), strict F1 (F1), overall Relevance (R), and overall pipeline score (O). The best variant is highlighted in \textbf{bold}.}
\label{tab:token_decision}
\resizebox{\columnwidth}{!}{%
\begin{tabular}{lcccc}
\toprule
\textbf{Maximum Tokens} & \textbf{P} & \textbf{F1} & \textbf{R} & \textbf{O} \\
\midrule
100           & 0.56 & 0.34 & 0.33 & 0.33 \\
\(\mathbf{200}\) & \textbf{0.54} & \textbf{0.34} & \textbf{0.33} & \textbf{0.34} \\
300          & 0.51 & 0.30 & 0.33 & 0.32 \\
\bottomrule
\end{tabular}%
}
\end{table}

\section{Post-Generation Attribution Parameter Experiment}
\label{appendix:threshold_experiments}
Experiments began from the answers generated by \texttt{LLaMA-3.3-70B} with one-shot prompting and fixed-$k$ of 54 as a retrieval strategy. We then performed a grid search over the three similarity weights $(w_{1},w_{2},w_{3})$ and the attribution threshold $T$ to identify the combination that maximizes the overall pipeline score, i.e., achieving higher strict attribution F1-score without unduly sacrificing answer relevance. Here, $w_1$, $w_2$, and $w_3$ correspond to the weights assigned to lexical, fuzzy, and semantic similarity scores. Each weight was varied in \(\{0.1,0.2,\dots,1.0\}\) under the constraint \(w_{1}+w_{2}+w_{3}=1\), and thresholds \(T\in\{0.1,0.2,\dots,0.9\}\) were tested. We observed that very low thresholds (0.1–0.2) led to over-attribution (nearly every answer sentence is attributed with every retrieved sentence), whereas very high thresholds (0.7–0.9) caused under-attribution (rarely answer sentences are attributed with retrieved sentences). Table~\ref{tab:postgen_grid} summarizes the top 10 configurations by strict F1-score. The best‐performing setting was \(\{w_{1}=0.0,\,w_{2}=0.5,\,w_{3}=0.5\}\) with $T=0.5$, yielding a strict F1-score 0.27 and overall pipeline score 0.31. 

\end{document}